# A Modelling Approach Based on Fuzzy Agents

Alain-Jérôme Fougères[1,2]

[1] ESTA, School of Business & Engineering
[2] IRTES-M3M, University of Technology of Belfort-Montbéliard,
Belfort, 90000, France
*ajfougeres@esta-belfort.fr*
*alain-jerome.fougeres@utbm.fr*

**Abstract**
Modelling of complex systems is mainly based on the decomposition of these systems in autonomous elements, and the identification and definitio9n of possible interactions between these elements. For this, the agent-based approach is a modelling solution often proposed. Complexity can also be due to external events or internal to systems, whose main characteristics are uncertainty, imprecision, or whose perception is subjective (*i.e.* interpreted). Insofar as fuzzy logic provides a solution for modelling uncertainty, the concept of fuzzy agent can model both the complexity and uncertainty. This paper focuses on introducing the concept of fuzzy agent: a classical architecture of agent is redefined according to a fuzzy perspective. A pedagogical illustration of fuzzy agentification of a smart watering system is then proposed.

## 1. Introduction

Complex systems are often characterized by the distribution of their components, as well as the distribution of knowledge to the general activity of these components (calculations, problem solving, etc.). When a problem cannot be solved by the concomitant action of several components of a complex system, sensitive to its environment, this leads to a series of interactions, often non-deterministic, between its components. Then the agent paradigm offers an interesting solution for modelling and developing complex systems. In addition, in real situations of decision support or cooperation, knowledge used by the components of complex systems is often inaccurate, incomplete, subject to uncertain assessments. Then, agents can be effectively used to handle complex uncertain problems where global knowledge is distributed and shared by a number of agents of a complex system, aiming to achieve a consensual solution in a collaborative way. Using intelligent agents that implement collaborative and distributed activity of complex system, with fuzzy logic [1], is recommended due to the uncertain nature of the collaboration, distribution, interaction in cooperative problems solving. So, how to model these agents?

Software agents are autonomous and distributed entities that are able to develop tasks either by themselves or by collaborating with other agents. An agent is a computer entity, located in an environment that it perceives, in which it acts [2]. This environment can be composed of other agents with whom it interacts independently. Fuzzy logic offers a framework for representing uncertainty and subjectivity of the real world. It has similarities with the way actors act as it uses a model of approximate reasoning that allows it to deal with design uncertainties. Agents, that implement uncertain problems by means of fuzzy logic, are called fuzzy agents [3, 4, 5]. So, how to apply the properties of agents to fuzzy agents?

In various fields of engineering (manufacturing, mobile robots, ambient intelligence, etc.), fuzzy agents was proposed as tool to model fuzzy behaviour problems, where agents can decide to act according to a fuzzy-logic rule base [6, 7, 8]. These agents can then use these rules to build a strategy for making decisions [9]. Fuzzy agents are also used in fuzzy reasoning situations, where agents interpret a situation, solve a problem or decide with fuzzy knowledge [4, 5, 7, 8]. Simulation of social relationships is also an area for experimentation fuzzy agents [10, 11]. Implementations of fuzzy agents are also proposed to solve distributed fuzzy problems [12], or improving the processing of the fuzziness of information, knowledge and interactions, in problem solving processes [13, 14, 3]. Despite the emergence of all this research, agents are not formalized enough to support the holistic view of tasks and processes performed by complex system. Given the need to design agents adapted to model, simulate and solve distributed problems with level of uncertainty, we proposed to better define the concept of fuzzy agents [13, 15]. This paper explains and develops the model of fuzzy agents, their interactions and their organizations.

The remainder of this paper is organized as follows: in the second section, with the perspective of fuzzy agent modelling, the main characteristics of fuzzy logic and fuzzy reasoning are presented. In the third section, modelling of fuzzy agent is proposed. In the fourth section implementation and application of fuzzy agents for a smart watering system is presented. In the last section, the conclusion shows some perspectives and interest in the proposed approach.

## 2. Fuzzy concept: set, logic and reasoning

A classic set has elements that satisfy all of its specific properties. More formally, a subset $A$ of a reference set $X$ can be described from its characteristic function $\chi_A : X \rightarrow \{0,1\}$ as follows (1):

$$\chi_A(x) = \begin{cases} 1 & si\ x \in A \\ 0 & sinon \end{cases} \quad (1)$$

Thus, elements that do not satisfy all the properties of $X$ cannot belong to this set.

However many subsets cannot be defined by a specific property: the subset "Warm Temperatures" of set "Temperatures", for example. It is then necessary to introduce a generalization of the characteristic function of membership of a subset of $A$ denoted $\mu_A$ which associates to each element $x$ of $X$ a real value $\mu_A(x)$ in the interval $[0, 1]$. This membership function allows highlighting grades membership of elements of the set $X$ and the reference to define a fuzzy subset of $X$. The operations defined on regular subsets (equality, inclusion, union, intersection, complement, etc.) are also used for fuzzy subsets. Thus, for a fuzzy relation $R$ between two universes of reference $X$ and $Y$ is defined membership function (2):

$$f_R : X \times Y \rightarrow [0,1] \quad (2)$$

Fuzzy logic, in its strict logical knowledge representation using fuzzy sets, is based on fuzzy elementary propositions as "$V$ is $A$" defined from a set $L$ of linguistic variables ($V$, $X$, $T_V$) where $V$ is a variable, $X$ is the universe in which it takes its values, and $T_V$ is a list of characterizations of $V$ represented by fuzzy subsets of $X$ [1].

For instance, let us consider the temperature $V$, and the list $T_V = \{A_1, A_2, A_3, A_4, A_5\}$, where $A_1$ = cold, $A_2$ = mild, $A_3$ = normal, $A_4$ = hot, and $A_5$ = burning (Fig. 5).

A fuzzy elementary proposition is constructed from a fuzzy subset $A$ of $T_V$ or a modified form of this fuzzy subset (weakening or strengthening). Its truth value belongs to any set $[0, 1]$, and it is provided by the membership function of the fuzzy set $\mu_A$. The proposition is especially true for any value $x$ of $X$ that $\mu_A(x)$ is high, so that $x$ is strongly characteristic $A$ of $V$. A truth value equal to 1 (respectively 0) corresponds to a proposition absolutely true (respectively absolutely false).

Fuzzy logic was introduced by Zadeh [1] as a framework for approximate reasoning. The fuzzy deductive reasoning can be considered an extension of the reasoning in classical logic. The basic operators of classical logic are also defined for the fuzzy logic:

- The conjunction of two fuzzy proposals "$V$ is $A$" and "$W$ is $B$" is a fuzzy proposal whose truth value $\mu_{A \wedge B}$ is obtained by aggregation using a *t-norm* of truth values of the two proposals (usually, $\mu_{A \wedge B}(x) = min(\mu_A(x), \mu_B(x))$).
- The disjunction of two fuzzy proposals "$V$ is $A$" and "$W$ is $B$" is a fuzzy proposal whose truth value $\mu_{A \vee B}$ is obtained by aggregation using a *t-conorm* truth values of the two proposals (usually, $\mu_{A \vee B}(x) = max(\mu_A(x), \mu_B(x))$).
- The negation of a fuzzy proposal "$V$ is $A$" is a fuzzy proposal whose truth value is $\neg \mu_A$ (usually, $\neg \mu_A(x) = 1 - \mu_A(x)$).
- The fuzzy implication is used to represent knowledge about a system in the form of rules such as "*IF $V$ is $A$ THEN $W$ is $B$*", built from linguistic variables ($V$, $X$, $T_V$) and ($W$, $Y$, $T_W$). Such a rule defines on $X \times Y$ a relation $R$, that is noted $A \rightarrow B$, between the values taken by $V$ and those taken by $W$. The relationship $A \rightarrow B$ determines the bonding strength between the premise "$V$ is $A$" and the conclusion "$W$ is $B$". Its membership function $\mu_{A \rightarrow B}$ corresponds to the truth value of the fuzzy implication between the two proposals. There are many forms of fuzzy implication which generally come from work on multivalued logic [16]. Thus, one proposed by Gödel-Brouwer (3) or that of Mamdani (4) where implication is seen as a conjunctive relation:

$$\mu_{A \rightarrow B}(x,y) = \begin{cases} 1 & si\ x \leq y \\ y & sin\ on \end{cases} \quad (3)$$

$$\mu_{A \rightarrow B}(x,y) = min(\mu_A(x), \mu_B(y)) \quad (4)$$

Extending the modus ponens, which allows to deduce that $q$ is true from the knowledge of a rule "*IF $p$ THEN $q$*" and the truth of $p$, Zadeh introduced the principle of compositional rule of inference, which is deduced easily from the principle of combination-projection. This rule allows to deduce information on the variable $W$ from the knowledge of a rule "*IF $V$ is $A$ THEN $W$ is $B$*" and a proposition "$V$ is $A'$" which should imperfectly to premise "$V$ is $A$" of the rule. The description $B'$ of $W$ which is obtained is defined by the membership function (5):

$$\mu_{B'}(y) = sup_{x \in X}\ F(\mu_{A'}(x), \mu_R(x,y)) \quad (5)$$

This rule provides the generalized modus ponens when $R$ is a fuzzy implication ($A \rightarrow B$). $F$ is called generalized modus ponens operator. Thus, if the operator $F$ is expressed by the minimum function, the generalized modus ponens can be written as (6):

$$\mu_{B'}(y) = \max_{x \in X}(\min(\mu_{A'}(x), \mu_{A \rightarrow B}(x,y))) \quad (6)$$

## 3. Fuzzy agent

### 3.1 Model

An intelligent agent is a computer system that is capable of flexible autonomous action in order to achieve the goals it has set (designed objectives). Such an agent is always located in an environment: it receives input from environment and acts to change this environment [2]. For Wooldridge [17], an agent is a system that enjoys the following properties: *Autonomy, Reactivity, Pro-activeness, Social ability*.

For Jennings [2], two central arguments for agent-based software engineering can be expressed: 1) *The Adequacy Hypothesis*. Agent-oriented approaches can significantly enhance our ability to model, design and build complex, distributed software systems; 2) *The Establishment Hypothesis*. As well as being suitable for designing and building complex systems, the agent-oriented approach will succeed as a mainstream software engineering paradigm.

Ferber [18] provides the following definition of the organization: organization assumes that there is a set of entities forming a certain unity and whose various elements are subordinated to each other in an integral unit and a convergent activity. Therefore, an organization requires a certain order between entities possibly heterogeneous, which contributes to the coherence.

Our research [19, 20, 15] focused on modelling agents with strong interactive capabilities (communication, cooperation, etc.), which may be used as basic components for the design of complex systems. A complex system is "made of many components with many interactions" [21]. So design of complex systems includes: 1) distribution and autonomy of system components, and 2) a very accurate modelling of communicative and interactional levels of these components. The agent-based approach provides a level of abstraction suitable for this problem [22].

Autonomy of an agent is technically implemented by: 1) an independent process, 2) an individual memory (knowledge / data of agent), and 3), an ability to interact with other agents and environment (perception / reception, emission / action). Many agent structures known as "cognitive" are inspired by the cycle *<perceive, decide, act>*. However, our generic agent model [15] is rather inspired by Rasmussen's three-level operator model [23]: 1) *reflex-based behaviour*, 2) *rule-based behaviour*, and 3) *knowledge-based behaviour* with interpretation, decision and plan. Agents are both cognitive and reactive. Moreover, they have behaviours adapted to the tasks they perform: Reactive task is characterised by the cycle *<Observation, Execution>*; routine task is characterised by the cycle *<Observation, Interpretation, Association state/task, Procedure/rules, Execution>*; finally cognitive task is characterised by the cycle: *<Observation, Interpretation, Decision of task, Planning, Execution>*.

The motivation of this research is that more effective design decisions can be made by fuzzy agents when fuzzy design information is considered in a fuzzy interaction based process [24]. Also, an agent-based system $\tilde{M}_\alpha$ will be fuzzy if agents that compose it are fuzzy. This means that agents have fuzzy knowledge and fuzzy behaviours, their interactions are fuzzy, their roles are fuzzy, and the resulting organizations are also fuzzy:

- Agents are fuzzy, that means that their knowledge (including the rules they use) are fuzzy (*i.e.*, defined with a fuzzy value, or with membership degrees in fuzzy sets) and their behaviours are fuzzy. The behaviour of an agent depends on the fuzzy evaluation of its perceptions, actions and decisions:

(i) fuzzy perceptions made by a fuzzy agent $\tilde{\alpha}_i$, defined by the function $\Phi_{\tilde{\Pi}(\tilde{\alpha}_i)} : \tilde{\Sigma} \times \tilde{\Sigma}_{\tilde{M}_{\tilde{\alpha}_i}} \rightarrow \tilde{\Pi}_{\tilde{\alpha}_i}$, depend on both the fuzzy states of $\tilde{\alpha}_i$ and fuzzy states of the agent-based system $\tilde{M}_\alpha$ (interpretations of fuzzy perceptions made by agent $\tilde{\alpha}_i$, depend on its fuzzy perceptions and fuzzy knowledge);

(ii) fuzzy decisions taken by a fuzzy agent $\tilde{\alpha}_i$, defined by the function $\Phi_{\tilde{\Delta}(\tilde{\alpha}_i)} : \tilde{\Pi}_{\tilde{\alpha}_i} \times \tilde{\Sigma}_{\tilde{\alpha}_i} \rightarrow \tilde{\Delta}_{\tilde{\alpha}_i}$, depend on fuzzy interpretations made by $\tilde{\alpha}_i$ with its fuzzy knowledge;

(iii) fuzzy actions performed by a fuzzy agent $\tilde{\alpha}_i$, defined by the function $\Phi_{\tilde{\Gamma}(\tilde{\alpha}_i)} : \tilde{\Delta}_{\tilde{\alpha}_i} \times \tilde{\Sigma} \rightarrow \tilde{\Gamma}_{\tilde{\alpha}_i}$, depend on the fuzzy decisions taken by $\tilde{\alpha}_i$ and fuzzy states of the agent-based system $\tilde{M}_\alpha$.

- Interactions between agents are fuzzy, since 1) the relationship (or affinities) between agents are weighted by a fuzzy value, and 2) interactions provide a relative interest (fuzzy evaluation) to agents based on roles that they play at a given time.

- Roles of agents are fuzzy, which means that all roles a fuzzy agent can play constantly have a fuzzy value. At a given time, it is possible to determine the roles that an agent play based on fuzzy values of its roles and a threshold value setting the minimum value an agent should invest in these roles.
- Organization (or organizations) of the agent-based system is fuzzy (and dynamic), insofar as the distribution of roles played by fuzzy agents is continually evolving – this defines a self-organizing agents which is the result both of their fuzzy multiple interactions and the continuing evolution of their roles in the global activity of the agent-based system.

A fuzzy agent-based system $\tilde{M}_\alpha$ is defined by (7):

$$\tilde{M}_\alpha = <\tilde{A}, \tilde{I}, \tilde{P}, \tilde{O}> \quad (7)$$

where $\tilde{A}$ is a fuzzy set of agents, $\tilde{I}$ is the fuzzy set of interactions between fuzzy agents of $\tilde{A}$, $\tilde{P}$ is the fuzzy set of roles that fuzzy agents of $\tilde{A}$ can play, and $\tilde{O}$ is the fuzzy set of organizations (or communities) defined for fuzzy agents of $\tilde{A}$. We can then affirm the flexibility of these organizations. However this flexibility is greater in matrix organizations than in hierarchical organizations.

In most agent-based systems, the behaviour of an agent that interacts with other agents of the system is composed of three phases (Fig. 1): 1) receive information from another agent or perceive a change in its environment (Fig. 2.a), 2) interpret this event and decide on actions to be performed taking into account other agents (Fig. 2.b), 3) send a message or perform an action modifying the environment (Fig. 2.c).

Thus, a fuzzy agent $\tilde{\alpha}_i \in \tilde{A}$ is described by (8):

$$\tilde{\alpha}_i = <\Phi_{\tilde{\Pi}(\tilde{\alpha}_i)}, \Phi_{\tilde{\Delta}(\tilde{\alpha}_i)}, \Phi_{\tilde{\Gamma}(\tilde{\alpha}_i)}, \tilde{K}_{\tilde{\alpha}_i}> \quad (8)$$

where $\Phi_{\tilde{\Pi}(\tilde{\alpha}_i)}$ is the function of fuzzy observations/perceptions of fuzzy agent $\tilde{\alpha}_i$; $\Phi_{\tilde{\Delta}(\tilde{\alpha}_i)}$ is the function of fuzzy decisions of fuzzy agent $\tilde{\alpha}_i$; $\Phi_{\tilde{\Gamma}(\tilde{\alpha}_i)}$ is the function of fuzzy actions of fuzzy agent $\tilde{\alpha}_i$; $\tilde{K}_{\tilde{\alpha}_i}$ is the finite set of fuzzy knowledge of fuzzy agent $\tilde{\alpha}_i$ - the knowledge contained in its memory, among which are the decision rules, the values of the domain, and the acquaintances and/or networks of affinities between agents, along with dynamic knowledge (observed events, internal states, etc.). The resource management associated with these various functions is provided by the set $\Psi$ of managers: $\Psi = \{\Psi_{\tilde{H}}, \Psi_{\tilde{\Gamma}}, \Psi_{\tilde{K}}\}$, where $\Psi_{\tilde{H}}$ is the messages manager, $\Psi_{\tilde{\Gamma}}$ is the actions manager and $\Psi_{\tilde{K}}$ is the knowledge-base manager (Fig. 1).

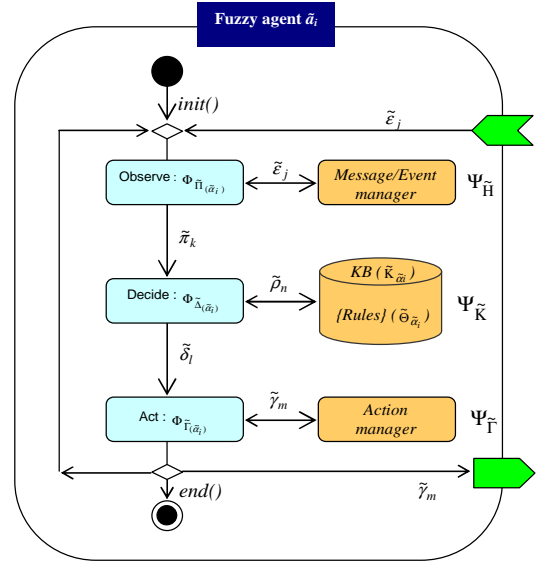

Fig. 1. Functional architecture of a fuzzy agent $\tilde{\alpha}_i$

3.2 Interactions between fuzzy agents

The literal definition of an interaction is « a reciprocal action of two or more phenomena ». In agent-based systems, as in human organizations, actions, interactions and communications (Fig. 3), are closely linked and interdependent [25]. Interaction is an exchange between agents and their environment. This exchange depends on the intrinsic properties of the world in which agents are active. The perception of agents may be passive when receiving messages / signals, or active, when it is the result of voluntary actions. Communication is an exchange between the agents themselves, using a language [26].

Communication in an agent-based system can be performed in two modes: 1) addressed communication to which a sender agent sends a message to one or more agents recipients (which corresponds to the model of Shannon), the basic unit in this communication is the speech act [27]; 2) unaddressed communication in which a sender agent sends a message to all agents available to the applicant in the environment (without recipients named).

If the interactions between agents are frequently communicative, they involve cooperation and coordination of actions. The agent-oriented coordination models focus on the behaviour of agents in order to achieve a coordinated system [28, 29]. El-Fallah Seghrouchni [30] classifies approaches to coordination in agent-based systems into six categories: 1) distributed problem solving, 2) organizational structure, 3) protocols, 4) negotiation, 5) formation of coalitions, and 6) planning.

A fuzzy interaction $\tilde{\iota}_i \in \tilde{I}$ between two fuzzy agents $\tilde{\alpha}_s$ and $\tilde{\alpha}_r$ is defined by the following tuple (9):

$$\tilde{\iota}_i = <\tilde{\alpha}_s, \tilde{\alpha}_r, \tilde{\gamma}_c> \qquad (9)$$

where $\tilde{\alpha}_s$ is the fuzzy agent source of the fuzzy interaction, $\tilde{\alpha}_r$ is the fuzzy agent destination of the fuzzy interaction, and $\tilde{\gamma}_c$ is a fuzzy act of cooperation. This cooperative act is consistent with the model of *5Co* we defined [22]: it belongs to the set *{Communication, Coordination, Co-production, Co-memory, Control-Process}*, and has a goal. Interactions are fuzzy; also a target fuzzy agent always evaluates an interaction (fuzzy value) to determine interest this interaction can take for it.

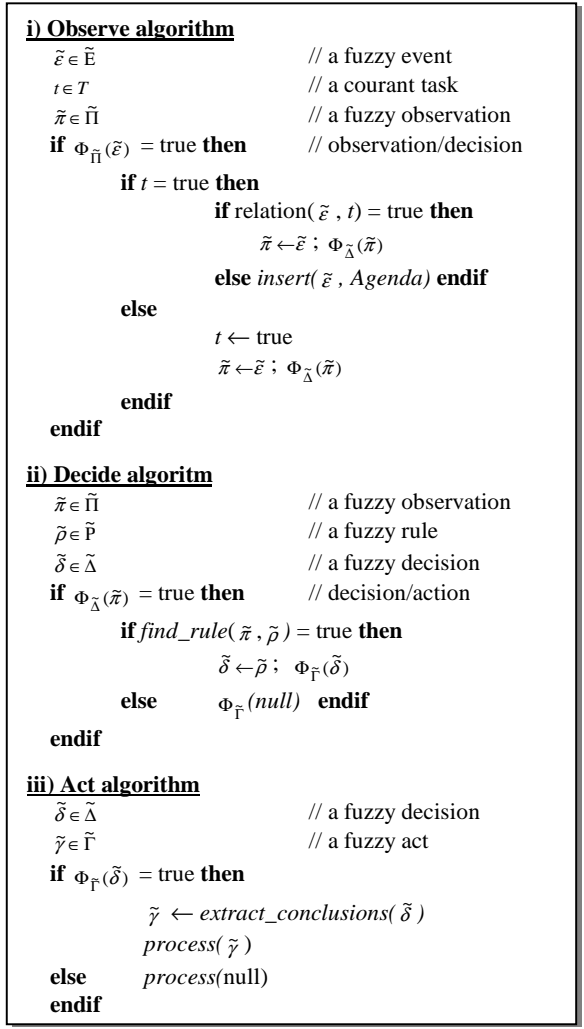

Fig. 2. Behavioural functions of fuzzy agents

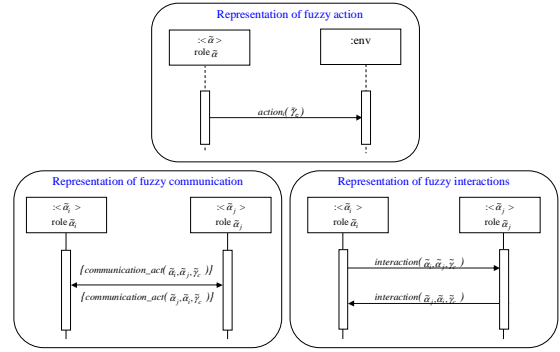

Fig. 3. Diagrammatic representation of concepts of fuzzy action, fuzzy interactions and fuzzy communication

In order to cooperate (*i.e.*, share information, ask services, negotiate fuzzy values, etc.), agents express their intention according to a language derived from the speech acts theory [27]. At least, fuzzy agents perform five speech acts: $\tilde{\Lambda} = \{inform, diffuse, ask, reply, confirm\}$ [14]. These five speech acts are sufficient to enable fuzzy agents to perceive intention of cooperation associated with the proposal contained in a fuzzy message. Cooperation is controlled by a protocol in which a response is required for some interactions (Fig. 4). For an interaction, a fuzzy agent chooses its fuzzy destination agent according to its intention, the context of configuration-solving and the state of its acquaintances. A fuzzy communication act $\tilde{\lambda}_{s,r}$ between two fuzzy agents ($\tilde{\lambda}_{s,r} \in \tilde{\Gamma}_{\tilde{\Lambda}\tilde{\alpha}_i}$) is defined by (10):

$$\tilde{\lambda}_{s,r} = <\tilde{\lambda}, \tilde{\alpha}_s, \tilde{\alpha}_r, \tau, \tilde{\eta}> \qquad (10)$$

which can be rewritten $\tilde{\iota}_i = <\tilde{\alpha}_s, \tilde{\alpha}_r, \tilde{\gamma}_{c_\lambda}>$ and $\tilde{\gamma}_{c_\lambda} = <\tilde{\lambda}_{s,r}, \tau, \tilde{\eta}>$, and where $\tilde{\lambda} \in \tilde{\Lambda}$ is a fuzzy speech act denoted by a performative verb, $\tilde{\alpha}_s$ is the fuzzy source agent of communication, $\tilde{\alpha}_r$ is the fuzzy receiver agent, $\tau \in T$ is a type of message (type or object), and $\tilde{\eta} \in \tilde{H}$ is the fuzzy message, which can be an assertion, a question, a response, etc. For instance, considering the speech act "*inform*" ($\tilde{\lambda}_{s,r} : \inf orm(\tilde{\alpha}_s(\mu_{\tilde{\alpha}_s}), \tilde{\alpha}_r(\mu_{\tilde{\alpha}_r}), \tau(1), \tilde{\eta}(\mu_{\tilde{\eta}}))$), we could obtain the following fuzzy value:
$\min(\tilde{\alpha}_s(\mu_{\tilde{\alpha}_s}) = 0.7, \tilde{\alpha}_r(\mu_{\tilde{\alpha}_r}) = 0.6, \tau(1) = 1, \tilde{\eta}(\mu_{\tilde{\eta}}) + 0.4) = 0.4$

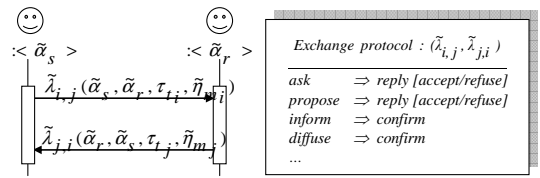

Fig. 4. Communication protocol between fuzzy agents

Each fuzzy agent plays a preset role, delimited by its fuzzy competences, and formalized by its fuzzy decision rules. Fuzzy decision rules $\tilde{\Delta}_{\tilde{\alpha}_i}$ of fuzzy agent $\tilde{\alpha}_i$, gathered in its knowledge base, are described by (11):

$$\tilde{\Delta}_{\tilde{\alpha}_i} = <\tilde{E}_{\tilde{\alpha}_i}, \tilde{X}_{\tilde{\alpha}_i}, \tilde{\Gamma}_{\tilde{\alpha}_i}> \qquad (11)$$

where $\tilde{E}_{\tilde{\alpha}_i}$ is the set of fuzzy events that fuzzy agent $\tilde{\alpha}_i$ can observe, $\tilde{X}_{\tilde{\alpha}_i}$ is the set of fuzzy conditions associated to the internal states of fuzzy agent $\tilde{\alpha}_i$, and $\tilde{\Gamma}_{\tilde{\alpha}_i}$ is the set of fuzzy actions that fuzzy agent $\tilde{\alpha}_i$ can perform.

For instance, let us consider the decision rule $\tilde{\delta}_i$ with: (1) $\tilde{\varepsilon}_1 := <inform, \tilde{\alpha}_s, \tilde{\alpha}_r, t = 2, V>$ ; (2) $\tilde{\chi}_1 := <V = sup(0.4)>$; (3) $\tilde{\gamma}_1 := <diffuse, \tilde{\alpha}_r, \tilde{A}_r, t=2, V>$.

This rule means that: (1) depending on following fuzzy event $\tilde{\varepsilon}_1$: the fuzzy agent $\tilde{\alpha}_r$ receives a message of type $t$ whose value is equal to 2 (corresponding to the transmission of a value) by which a fuzzy agent $\tilde{\alpha}_s$ informs $\tilde{\alpha}_r$ of its value $V$; (2) under condition $\tilde{\chi}_1$ "$V$ must be greater than the threshold value 0.4"; (3) action $\tilde{\gamma}_1$ will then be triggered: agent $\tilde{\alpha}_r$ will communicate this information to all agents of its community $\tilde{A}_r$. Actions of each agent $\tilde{\alpha}_i$ are controlled and memorized by a manager $M_{\tilde{\Gamma}_{\tilde{\alpha}_i}}$.

The rule $\tilde{\delta}_i$: *IF $\tilde{\varepsilon}_1$ AND $\tilde{\chi}_1$ THEN $\tilde{\gamma}_1$* will be triggered by the fuzzy agent $\tilde{\alpha}_r$, depending on the value of the premises and a threshold, the *AND* of the fuzzy rule can be defined by the operator *MIN* (for instance, $MIN(\tilde{\varepsilon}_1, \tilde{\chi}_1) \geq 0.4$).

### 3.3 Organization in fuzzy agent-based systems

The problems inherent in the partial knowledge of agents (pursuit of local goals, interleaving activities, etc.), require the development of elaborate coordination mechanisms [32]. The organization shall allow an agent-based system to behave as a coherent whole, to solve a problem unequivocally. It controls and coordinates the interactions between agents of the system, thus structuring their activities with the goal of convergence. Ferber and al. [18] distinguish between "organizational structure" and "organization", corresponding to the process of designing the structure.

Two definitions complement the previous: 1) Gasser [33] which proposed an organization "provides a framework for activity and interaction through the definition of roles, behavioural expectations and authority relationships"; and 2) Wooldridge [34] which proposed a more practical definition, where an organisation is viewed as "a collection of roles, that stand in certain relationships to one another, and that take part in systematic institutionalised patterns of interactions with other roles".

From these basic definitions, we extract the following characteristics:
- In an agent-based system, an organization consists of agents, active entity whose behaviour and well-defined functionalities.
- An organization can be partitioned into groups or communities of agents.
- A group (or community) is comprised of agents sharing a goal and characteristics.
- An agent can belong to several groups.
- An agent can play one or more roles within the group or groups to which he belongs.
- An agent interacts with agents of its community or other communities to carry out the roles it should play.
- A role is an abstract representation of a function to be performed by an agent in a group.

In a dynamic organizational structure, the roles of fuzzy agents can become dynamic, variable and determined by actions to be performed. We proposed that the roles of fuzzy agents are considered fuzzy, and defined as follows (12):

$$\tilde{P} = \{\tilde{\rho}_1, \tilde{\rho}_2, ..., \tilde{\rho}_q\} \qquad (12)$$

Then, the fuzzy set of roles played by an agent $\tilde{\alpha}_i$ is defined by (13):

$$\tilde{P}(\tilde{\alpha}_i) = \{\mu_{\tilde{\rho}_1}(\tilde{\alpha}_i), \mu_{\tilde{\rho}_2}(\tilde{\alpha}_i), ..., \mu_{\tilde{\rho}_q}(\tilde{\alpha}_i)\} \qquad (13)$$

Let $\Phi_{\tilde{P}} : \tilde{A} \to \{\tilde{P}\}$ be the function "play a role", then the roles $\tilde{\rho}_j \in \tilde{P}$ played by a fuzzy agent $\tilde{\alpha}_i$ are defined by $\Phi_{\tilde{P}}(\tilde{\alpha}_i, \tilde{\rho}_j) = \{\mu_{\tilde{\alpha}_i}(\tilde{\rho}_j)\}$, with $j \in J_{\tilde{P}}$, $J_{\tilde{P}} = \{1, 2, ..., q_{\tilde{P}}\}$.

A fuzzy agent-based system can be divided/organized into communities as follow (14):

$$\tilde{A}_1 \subseteq \tilde{A}, ..., \tilde{A}_i \subseteq \tilde{A}, \quad \text{where } n \text{ is the number of communities} \qquad (14)$$

Each community has a clear objective, which determines the main role that fuzzy agents will play in this community. This means that each fuzzy agent belongs to a community of reference in which it plays its main role (15):

$$\forall \tilde{\alpha} \in \tilde{A} \supset [\exists x, \tilde{\alpha} \in \tilde{A}_x \wedge \Phi_{\tilde{P}}(\tilde{\alpha}, \tilde{\rho}_x)] \quad (15)$$

Fuzzy agents interact by sending messages within their communities (in this case they emphasize participation in their main role), but they also interact with fuzzy agents from other communities (see table above), in which case they involved in other roles. A fuzzy agent $\tilde{\alpha}_i$ by interacting with a fuzzy agent $\tilde{\alpha}_j$ of another community then participates in the same role as $\tilde{\alpha}_j$ (16):

$$\forall \tilde{\alpha}_i \in \tilde{A} \supset [\exists x : \tilde{\rho}_x \in \tilde{P} \wedge \alpha_j \in \tilde{A}_x, \Phi_{\tilde{P}}(\tilde{\alpha}_j, \tilde{\rho}_x) \\ \wedge \tilde{\lambda}_{i,j}(\tilde{\alpha}_i, \tilde{\alpha}_j, \tau, \tilde{\eta}) \supset \Phi_{\tilde{P}}(\tilde{\alpha}_i, \tilde{\rho}_x)] \quad (16)$$

### 3.4 Fuzzy agentification methodology

First phase is to determine the universes $U_{1..n}$ of the domain or system considered. Then, agentification can be performed in two other phases:

1) Distribution of universes $U_{1..n}$. Each universe $U_i$ is discretized in order to establish a set of features characteristic of the universe - this phase allows adjusting the granularity of the future agent-based system. Each characteristic entity is then agentified: creation of a fuzzy agent for each entity (corresponding to a bijection).

2) Acquisition and representation of knowledge of fuzzy agents. Each fuzzy agent acts autonomously according to its own knowledge. These include: membership degrees to different fuzzy subsets defined in the universe in which it operates, fuzzy rules that define its behaviour, fuzzy rules that define relationships or interactions with other fuzzy agents.

## 4. Illustration: a smart watering system

To illustrate the approach, we chose a lawn watering system in fuzzy control. Placed on the ground, the system triggers from its start watering a garden lawn during a specified period (output parameter of the system) with input parameters are soil humidity and the outside temperature (*humidity* × *temperature* → *watering_duration*).

Three universes are defined: 1) Universe $U_1$ representing the temperature ranging from 0 to 45 degrees, 2) universe $U_2$ representing the humidity over an interval of 0 to 30% humidity, and 3) universe $U_3$ representing the watering duration, from 0 to 70 minutes. Fuzzy sets and membership functions are shown in the three following figures (Fig. 5, Fig. 6 and Fig. 7).

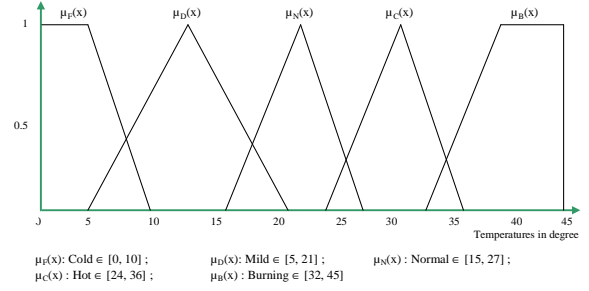

$\mu_F(x)$: Cold ∈ [0, 10] ; $\mu_D(x)$: Mild ∈ [5, 21] ; $\mu_N(x)$ : Normal ∈ [15, 27] ;
$\mu_C(x)$ : Hot ∈ [24, 36] ; $\mu_B(x)$ : Burning ∈ [32, 45]

Fig. 5. Membership functions for fuzzy subsets of "temperature"

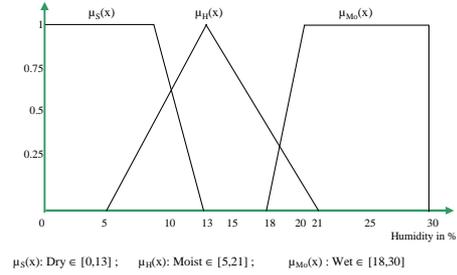

$\mu_S(x)$: Dry ∈ [0,13] ; $\mu_H(x)$: Moist ∈ [5,21] ; $\mu_{Mo}(x)$ : Wet ∈ [18,30]

Fig. 6. Membership functions for fuzzy subsets "humidity"

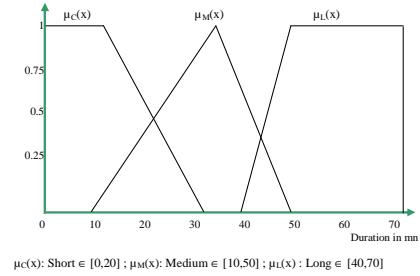

$\mu_C(x)$: Short ∈ [0,20] ; $\mu_M(x)$: Medium ∈ [10,50] ; $\mu_L(x)$ : Long ∈ [40,70]

Fig. 7. Membership functions for fuzzy subsets of "watering duration"

Each fuzzy agent of the watering system activates the fuzzy rules contained in its knowledge base whose premises are under fuzzy inputs from the fuzzification of real inputs acquired periodically. These rules are of the form shown in (17). The following illustrations are based on fuzzy rules of fuzzy agent <watering_duration> (18, 19, 20, 21):

$$\delta_x \tilde{\alpha}_y : \text{IF } \{event \wedge condition\} \text{ THEN } \{action\} \quad (17)$$

Let $\tilde{\alpha}_T$ the fuzzy agent representing the universe of temperatures, $\tilde{\alpha}_H$ the fuzzy agent representing the universe of humidity, and $\tilde{\alpha}_D$ the fuzzy agent representing the universe of watering duration.

$\delta_i \tilde{\alpha}_D$ : IF *temperature is burning* AND *humidity is moist* THEN *the watering duration is average* (18)

$\delta_j \tilde{\alpha}_D$ : IF *temperature is burning* AND *humidity is dry*
THEN *the watering duration is long* (19)

$\delta_k \tilde{\alpha}_T$ : IF $\mu'_B(\tilde{\alpha}_T) \neq \mu_B(\tilde{\alpha}_T)$ // temperature exchange "hot"
THEN $\inf orm(\tilde{\alpha}_T, \tilde{\alpha}_D, T, \mu'_B(\tilde{\alpha}_T))$ (20)

$\delta_l \tilde{\alpha}_H$ : IF $\mu'_S(\tilde{\alpha}_H) \neq \mu_S(\tilde{\alpha}_H)$ // humidity exchange "wet"
THEN $\inf orm(\tilde{\alpha}_H, \tilde{\alpha}_D, T, \mu'_S(\tilde{\alpha}_H))$ (21)

To illustrate the activation of the above rules, we consider the following scenario (Fig. 8): the fuzzy agent $\tilde{\alpha}_T$ observes that temperature changes (event $e_i$); it evaluates the new degree of membership to the fuzzy set "burning temperature" $\mu'_B(\tilde{\alpha}_T)$; then inform the fuzzy agent $\tilde{\alpha}_T$ (action $a_i$). The latter evaluates the degree of membership to the fuzzy set "medium watering," $\mu'_L(\tilde{\alpha}_D)$ and then adjusts watering (action $a_k$).

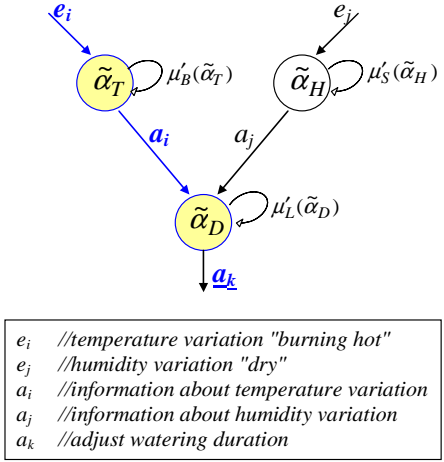

$e_i$ //temperature variation "burning hot"
$e_j$ //humidity variation "dry"
$a_i$ //information about temperature variation
$a_j$ //information about humidity variation
$a_k$ //adjust watering duration

Fig. 8. Typical interaction network

For instance, consider a new temperature of 35 ° and a constant humidity of 10% (Fig. 9). The fuzzy agent $\tilde{\alpha}_T$ observed the new temperature, evaluates new degree of membership to the fuzzy set "burning temperature" $\mu'_B(\tilde{\alpha}_T) = 0.45$, and then informs the fuzzy agent $\tilde{\alpha}_D$. Given the humidity to 10%, the fuzzy agent $\tilde{\alpha}_H$ had previously interacts with the fuzzy agent $\tilde{\alpha}_D$: information on the degree of membership to the fuzzy set "dry humidity" $\mu'_S(\tilde{\alpha}_H) = 0.35$, and information on the degree of membership to the fuzzy set "wet humidity" $\mu'_H(\tilde{\alpha}_H) = 0.61$. The fuzzy agent $\tilde{\alpha}_D$ then triggers the rules $\delta_i \tilde{\alpha}_D : \min(0.45, 0.61) = 0.45 \rightarrow moyenne$ and $\delta_j \tilde{\alpha}_D : \min(0.45, 0.35) = 0.35 \rightarrow longue$, then, conducting the defuzzification by the Centre of area method, obtains a new watering duration of 40 minutes.

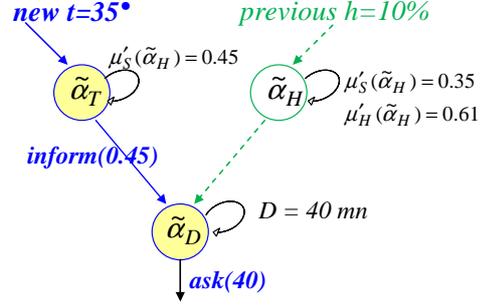

Fig. 9. Effective interaction network

## 5. Conclusion and Perspectives

In this paper, we presented a model of fuzzy agents proposed for the modelling and design of complex systems (intelligent/smart systems, distributed systems, cooperative systems, assistance systems, etc.), where uncertainty and imprecision are considered. The correlated formal approach is: a) to define a modular architecture for designing the various fuzzy cognitive processes of fuzzy agents, b) to respect a rigorous methodology to acquire the fuzzy expertise of each fuzzy agent, c) to define the fuzzy behaviour and the evolving of the fuzzy roles of each fuzzy agents, and d) to define the model of fuzzy knowledge and fuzzy interactions of each fuzzy agent. A simple and pedagogical case study of fuzzy agentification (a smart watering system) was presented to illustrate our approach.

We are now working on two ways: 1) a better understanding and modelling of role changing of fuzzy agents during their activities for cooperative problem solving, and 2) application of the model to cooperative information systems (mainly in the fields of uncertain information management and support for the relevance of collaborative work), collaborative design (mainly in the field of fuzzy product configuration), and natural language processing (mainly in the fields of understanding and formalization of functional descriptions).

## Appendix: notations used in the fuzzy agent model

$\tilde{A} = \{\tilde{\alpha}_i\}$ is the finite fuzzy set of fuzzy agents, $i \in I_{\tilde{A}}$, $I_{\tilde{A}} = \{1, 2, ..., q_{\tilde{A}}\}$ ;

$\tilde{I} = \{\tilde{\iota}_i\}$ is the finite fuzzy set of interactions defined for all fuzzy agents, $i \in I_{\tilde{I}}$, $I_{\tilde{I}} = \{1, 2, ..., q_{\tilde{I}}\}$ ;

$\tilde{P} = \{\tilde{\rho}_i\}$ is the finite fuzzy set of roles to be played by all fuzzy agents, $i \in I_{\tilde{P}}$, $I_{\tilde{P}} = \{1,2,...,q_{\tilde{P}}\}$ ;

$\tilde{O} = \{\tilde{o}_i\}$ is the finite fuzzy set of organizations of all fuzzy agents into communities, $i \in I_{\tilde{O}}$, $I_{\tilde{O}} = \{1,2,...,q_{\tilde{O}}\}$ ;

$\tilde{\Sigma} = \{\tilde{\sigma}_i\}$ is the finite fuzzy set of states of agent-based system, $i \in I_{\tilde{\Sigma}}$, $I_{\tilde{\Sigma}} = \{1,2,...,q_{\tilde{\Sigma}}\}$ ;

$\tilde{\Sigma}_{\tilde{\alpha}_i} \subseteq \tilde{\Sigma}$ is the finite fuzzy set of states of fuzzy agent $\tilde{\alpha}_i$ ;

$\tilde{\Sigma}_{\tilde{M}_{\tilde{\alpha}_i}} \subseteq \tilde{\Sigma}$ is the finite fuzzy set of states of agent-based system that fuzzy agent $\tilde{\alpha}_i$ knows;

$\tilde{\Pi} = \{\tilde{\pi}_i\}$ is the finite fuzzy set of perceptions, $i \in I_{\tilde{\pi}}$, $I_{\tilde{\pi}} = \{1,2,...,q_{\tilde{\pi}}\}$ ;

$\tilde{\Pi}_{\tilde{\alpha}_i} \subseteq \tilde{\Pi}$ is the finite fuzzy set of perceptions of fuzzy agent $\tilde{\alpha}_i$ ;

$\tilde{\Delta} = \{\tilde{\delta}_i\}$ is the finite fuzzy set of fuzzy decisions, $i \in I_{\tilde{\Delta}}$, $I_{\tilde{\Delta}} = \{1,2,...,q_{\tilde{\Delta}}\}$ ;

$\tilde{\Gamma} = \{\tilde{\gamma}_i\}$ is the finite fuzzy set of actions, $i \in I_{\tilde{\Gamma}}$, $I_{\tilde{\Gamma}} = \{1,2,...,q_{\tilde{\Gamma}}\}$ ;

$\tilde{\Gamma}_{\tilde{\alpha}_i} \subseteq \tilde{\Gamma}$ is the finite fuzzy set of actions that fuzzy agent $\tilde{\alpha}_i$ can process;

$\tilde{\Gamma}_{\tilde{\Lambda}_{\tilde{\alpha}_i}} \subseteq \tilde{\Gamma}$ is the specific finite fuzzy set of communication acts that fuzzy agent $\tilde{\alpha}_i$ can process;

$\tilde{\Omega} = \{\tilde{\omega}_i\}$ is the finite fuzzy set of reactions caused by actions of fuzzy agent $\tilde{\alpha}_i$ on its environment, $i \in I_{\tilde{\Omega}}$, $I_{\tilde{\Omega}} = \{1,2,...,q_{\tilde{\Omega}}\}$ ;

$\tilde{K} = \{\tilde{\kappa}_i\}$ is the finite fuzzy set of fuzzy knowledge, $i \in I_{\tilde{K}}$, $I_{\tilde{K}} = \{1,2,...,q_{\tilde{K}}\}$ ;

$\tilde{K}_{\tilde{\alpha}_i} \subseteq \tilde{K}$ is the finite fuzzy set of fuzzy knowledge of fuzzy agent $\tilde{\alpha}_i$, with $\tilde{K}_{\tilde{\alpha}_i} = \tilde{P}_{\tilde{\alpha}_i} \cup \tilde{\Sigma}_{\tilde{\alpha}_i} \cup \tilde{\Sigma}_{\tilde{M}_{\tilde{\alpha}_i}}$ ;

$\tilde{E} = \{\tilde{\varepsilon}_i\}$ is the finite fuzzy set of fuzzy events, $i \in I_{\tilde{E}}$, $I_{\tilde{E}} = \{1,2,...,q_{\tilde{E}}\}$ ;

$\tilde{E}_{\tilde{\alpha}_i} \subseteq \tilde{E}$ is the finite fuzzy set of fuzzy events that fuzzy agent $\tilde{\alpha}_i$ can observe ;

$\tilde{X} = \{\tilde{\chi}_i\}$ is the finite fuzzy set of conditions, $i \in I_{\tilde{X}}$, $I_{\tilde{X}} = \{1,2,...,q_{\tilde{X}}\}$ ;

$\tilde{X}_{\tilde{\alpha}_i} \in \tilde{X}$ is the finite fuzzy set of conditions associated to the internal states of fuzzy agent $\tilde{\alpha}_i$ ;

$\tilde{N} = \{\tilde{v}_i\}$ is the finite fuzzy set of configuration networks, $i \in I_{\tilde{N}}$, $I_{\tilde{N}} = \{1,2,...,q_{\tilde{N}}\}$ ;

$\tilde{Y} = \{\tilde{v}_i\}$ is the finite fuzzy set of connexions between fuzzy solution agents, $i \in I_{\tilde{Y}}$, $I_{\tilde{Y}} = \{1,2,...,q_{\tilde{Y}}\}$ ;

$\tilde{B} = \{\tilde{\beta}_i\}$ is the finite fuzzy set of speech acts, $i \in I_{\tilde{B}}$, $I_{\tilde{B}} = \{1,2,...,q_{\tilde{B}}\}$ .

$\tilde{H} = \{\tilde{\eta}_i\}$ is the finite fuzzy set of messages, $i \in I_{\tilde{H}}$, $I_{\tilde{H}} = \{1,2,...q_{\tilde{H}}\}$ ;

$\tilde{T} = \{\tilde{\tau}_i\}$ is the finite set of type of messages, $i \in I_{\tilde{T}}$, $I_{\tilde{T}} = \{1,2,...,q_{\tilde{T}}\}$ ;

$\Phi_{\tilde{\Pi}(\tilde{\alpha}_i)} : \tilde{\Sigma} \times \tilde{\Sigma}_{\tilde{M}_{\tilde{\alpha}_i}} \to \tilde{\Pi}_{\tilde{\alpha}_i}$ is the function of observations of fuzzy agent $\tilde{\alpha}_i$ ;

$\Phi_{\tilde{\Delta}(\tilde{\alpha}_i)} : \tilde{\Pi}_{\tilde{\alpha}_i} \times \tilde{\Sigma}_{\tilde{\alpha}_i} \to \tilde{\Delta}_{\tilde{\alpha}_i}$ is the function of decisions of fuzzy agent $\tilde{\alpha}_i$ ;

$\Phi_{\tilde{\Gamma}(\tilde{\alpha}_i)} : \tilde{\Delta}_{\tilde{\alpha}_i} \times \tilde{\Sigma} \to \tilde{\Gamma}_{\tilde{\alpha}_i}$ is the function of actions of fuzzy agent $\tilde{\alpha}_i$ .

## Acknowledgments

We would like to warmly thank Dr. Egon Ostrosi and Dr. Denis Choulier from Laboratory IRTES-M3M for the fruitful discussions on the paradigm of "fuzzy agent", and for the opportunity they give us to implement our research on cooperative design of mechanical systems.

**Alain-Jérôme Fougères** is holder of a PhD in Artificial Intelligence from the University of Technology of Compiègne - France (1997). He is Professor of Computer Science at ESTA and he conducts his research at ESTA-Lab. He is also associated to the IRTES institute of University of Technology of Belfort-Montbéliard. Initially, his areas of interests and his scientific contributions were: 1) assistance to the redaction of formal specifications, where problems focused on the natural language processing, the knowledge representation, and the techniques of formal specifications; and 2) design of agent-based systems, in particular architecture, interactions, communication, and co-operation problems. The last ten years his research focuses on agent-based mediation for co-operative work.